\documentclass[10pt,twocolumn,letterpaper]{article}
\usepackage[]{wacv}      
\usepackage{booktabs}
\usepackage{multirow}
\usepackage{amssymb}
\usepackage{pifont}  
\usepackage{makecell}
\definecolor{wacvblue}{rgb}{0.21,0.49,0.74}
\usepackage[pagebackref,breaklinks,colorlinks,allcolors=wacvblue]{hyperref}
\usepackage[capitalize]{cleveref}
\crefname{section}{Sec.}{Secs.}
\Crefname{section}{Section}{Sections}
\Crefname{table}{Table}{Tables}
\crefname{table}{Tab.}{Tabs.}
\def \ie {\emph{i.e.},}
\def \eg {\emph{e.g.},}
\def \etal {\emph{et al.}}

\newcommand{\tit}[1]{\smallbreak\noindent\textbf{#1.}}

\newcommand{\xmark}{\ding{55}}%

\title{Autoregressive Styled Text Image Generation, but Make it Reliable}

\author{
Carmine Zaccagnino$^{1}$ \quad Fabio Quattrini$^{1}$ \quad Vittorio Pippi$^{1}$ \\ \quad Silvia Cascianelli$^1$ \quad Alessio Tonioni$^{2}$ \quad Rita Cucchiara$^1$\\
\begin{tabular}{cc}
\makecell{$^1$University of Modena and Reggio Emilia} & \makecell{$^2$Google}
\end{tabular}\\
\url{https://aimagelab.github.io/Eruku}
}

\begin{document}
\maketitle

\begin{abstract}
Generating faithful and readable styled text images (especially for Styled Handwritten Text generation - HTG) is an open problem with several possible applications across graphic design, document understanding, and image editing.
A lot of research effort in this task is dedicated to developing strategies that reproduce the stylistic characteristics of a given writer, with promising results in terms of style fidelity and generalization achieved by the recently proposed Autoregressive Transformer paradigm for HTG. However, this method requires additional inputs, lacks a proper stop mechanism, and might end up in repetition loops, generating visual artifacts. In this work, we rethink the autoregressive formulation by framing HTG as a multimodal prompt-conditioned generation task, and tackle the content controllability issues by introducing special textual input tokens for better alignment with the visual ones. Moreover, we devise a Classifier-Free-Guidance-based strategy for our autoregressive model. Through extensive experimental validation, we demonstrate that our approach, dubbed \textbf{Eruku}, compared to previous solutions requires fewer inputs,  generalizes better to unseen styles, and follows more faithfully the textual prompt, improving content adherence.
\end{abstract}
    
\section{Introduction}
\label{sec:intro}

Generating images containing some desired string in a specific style is a challenging task that has drawn renewed interest in the Computer Vision and Document Analysis communities~\cite{yousef2020origaminet,cascianelli2021learning,cascianelli2022boosting,quattrini2024mu}. While state-of-the-art generative models, with notoriously poor performance on this task, are making steady progress for generic and simple styles~\cite{liu2024glyph,liu2024glyphv2,tuo2023anytext,tuo2024anytext2}, they still are not being applied to the details-oriented variant that focuses on handwriting, \ie~Handwritten Text Generation (HTG)~\cite{graves2013generating,haines2016my,alonso2019adversarial}.
Typically, models for HTG take as input one or more style images, containing text written in a reference handwriting style, and a text string that specifies some desired content. Then, the models are tasked to generate another image containing the desired textual content in the reference style.

Research efforts in the last few years have brought to impressive performance with models following mainly the adversarial (GAN-based)~\cite{kang2020ganwriting,gan2022higan+,vanherle2024vatr++} or the diffusion-based~\cite{nikolaidou2023wordstylist,dai2024one,nikolaidou2024diffusionpen} generative paradigms. However, these kinds of approaches exhibit poor generalization capabilities when tasked to generate images in handwriting styles that differ substantially from those observed during training~\cite{pippi2023choose,pippi2025quo}. Moreover, they typically impose constraints on the output length, and are difficult to train, usually requiring multiple auxiliary networks or supervision signals to ensure style fidelity and content readability in the output.

\begin{figure}
    \centering
    \includegraphics[width=\linewidth]{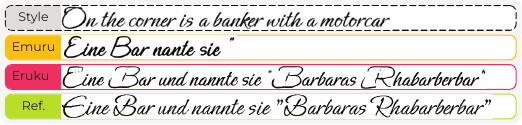}
    \caption{Our proposed Eruku model can generate text images with arbitrary length and with great text adherence, respecting both the generation text and with the conditioning writing style.}\vspace{-0.5cm}
    \label{fig:overview}
\end{figure}

More recently, Pippi~\etal~\cite{pippi2025zero} tackled these issues by proposing to treat HTG as an autoregressive image generation problem. 
Specifically, their model is given a text line image as reference style example, alongside its transcription. The image is represented as a sequence of visual embeddings obtained with a Variational AutoEncoder (VAE). Then, the model autoregressively generates the VAE-compatible visual embeddings of an image containing the desired text in the style of the reference example. 
This approach generalizes well to novel styles, both handwritten and typewritten, thanks to training on massive synthetic datasets. Moreover, it does not have architectural restrictions preventing it from generating arbitrarily long images, and is trained with a simple loss that does not entail terms from external models.
However, the approach proposed in~\cite{pippi2025zero} also presents important drawbacks. First, it requires as input the transcription of the style image. This helps the model associate style features with textual content, but creates a strong dependency on accurate transcriptions, which may not be available in real-world scenarios, or may be unreliable when obtained with text recognition networks, which can be imperfect. Moreover, to stop the generation, it relies on a heuristic that entails emitting 10 consecutive padding tokens, which are then discarded. This strategy is somewhat inefficient. 
Finally, the model often struggles to precisely render the desired text, suffering from issues typical of autoregressive generation models such as repetitions, incomplete sequences, and failure to stop at the correct length (see~\Cref{fig:overview}).

Nevertheless, the autoregressive formulation provides undeniable advantages like training efficiency and the ability to generate arbitrary-length outputs. Therefore, in this work, we follow the same formulation, improving its key aspects. 
To this end, we introduce modeling novelties to
\begin{itemize}
\item free the model from requiring the transcription of the style image as input, making it instead optional;
\item provide the model with an explicit stopping mechanism via a single, dedicated end of generation token;
\item enforce adherence to the desired text sequence without relying on auxiliary networks for supervision.
\end{itemize}
We achieve these goals by introducing special visual and textual tokens to guide the generation, and by a Classifier-Free Guidance (CFG)-inspired approach that works only on the textual inputs.

Our method, dubbed \textit{Eruku}, is built upon a VAE and an autoregressive Transformer trained on a large-scale synthetic dataset of text images. Specifically, the Transformer is trained to iteratively predict VAE-compatible embeddings, including our introduced special tokens, and to generate by exploiting our CFG-inspired mechanism.

We conduct extensive experiments on multiple handwritten and typewritten datasets, all different from the synthetic one used in training. The obtained results show that Eruku achieves robustness to missing or noisy inputs and improves text fidelity while maintaining strong generalization to unseen styles. The code and weights of our approach will be available upon publication.

\section{Related Work}
\label{sec:related}
\begin{figure*}[t]
    \centering
    \includegraphics[width=\linewidth]{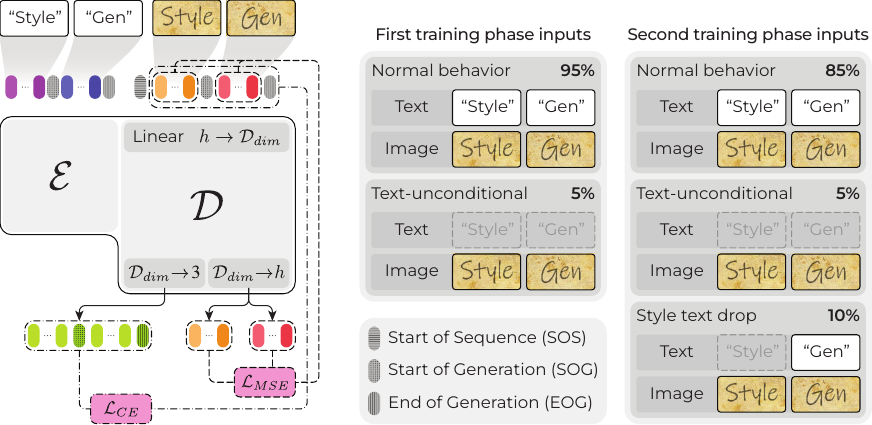}
    \caption{Training framework of Eruku, our autoregressive text image generation model. We condition generation on: the textual content of a style image $T_s$ (\textit{``Style''}), the generation text $T_g$ (\textit{``Gen''}), and the style image $I_s$. Eruku is trained on next-token prediction, learning to generate an image containing the generation text $T_g$ with the same writing style as the style image $I_s$. Providing the style text $T_s$ enables the model to link each character with its representation style, but we also enable generation without the style text $T_s$ by dropping it during inference and using synchronization tokens to separate the sequence components. We represent images with VAE~\cite{pippi2025zero} continuous latents. Our model automatically learns to stop generation emitting the \textit{Visual End of Generation token} $\texttt{<EOG>}$.}\vspace{-0.3cm}
    \label{fig:pipeline}
\end{figure*}

HTG approaches can be broadly categorized into two paradigms: \emph{online} and \emph{offline}. Online HTG conceptualizes handwriting as a temporal sequence of strokes. Models in this setting take as input a style representation encoded in the form of stroke trajectories and then predict subsequent trajectories, which are finally rendered as an image~\cite{graves2013generating,aksan2018deepwriting,aksan2018stcn,dai2023disentangling,luhman2020diffusion,ren2023diff}. However, the need for specialized hardware such as digitizing tablets to capture stroke-level data makes this approach costly and impractical in scenarios where the reference style is derived from existing manuscripts, such as historical collections. This limitation has motivated the development of offline HTG, where both the style reference and the generated samples are static images. Offline HTG has therefore attracted more research attention and also constitutes the focus of this paper.
Offline HTG aims at generating text images that reflect a user-defined content string~\cite{alonso2019adversarial,fogel2020scrabblegan} and, in the Styled variant focus of this work, one or more reference style images~\cite{kang2020ganwriting,bhunia2021handwriting,pippi2023handwritten,vanherle2024vatr++,davis2020text,pippi2025zero,gan2021higan,gan2022higan+,krishnan2023textstylebrush,dai2024one,mayr2024zero}. For our model, we use a single text line image as reference since this provides rich stylistic cues while remaining convenient for the user to supply.

Most HTG models are either GAN-based~\cite{kang2020ganwriting,davis2020text,gan2021higan,gan2022higan+,luo2022slogan,krishnan2023textstylebrush} and diffusion-based frameworks~\cite{zhu2023conditional,nikolaidou2023wordstylist,nikolaidou2024diffusionpen,dai2024one}, and rely on convolutional backbones. In these methods, content and style are encoded separately and fused in later stages, which prevents the modeling of content-sensitive stylistic phenomena (\eg~ligatures or character-specific rendering variations such as repeated letters). To address this issue, recent Transformer-based adversarial approaches~\cite{bhunia2021handwriting,pippi2023handwritten,vanherle2024vatr++} introduce cross-attentions to capture style–content interactions. To further exploit such interactions,~\cite{pippi2025zero} introduces an encoder-decoder autoregressive Transformer that generates styled text images by conditioning on both the style image and its textual content, learning how to link the style characters with their visual appearance. This introduces two limitations: first, the required textual content of the style image may not always be available or easy to obtain; second, when the style is particularly complex, the binding between the style image and its text content can fail, and generation is prone to collapse.
To address this, we introduce explicit synchronization and stop tokens, which make Eruku able to generate styled text even when not provided with the textual content of the style image. 

The majority of existing HTG methods target short sequences, typically single words. As a consequence, when longer outputs are attempted, models often fail to preserve character quality, proportion, and consistency across the output image. Moreover, fixed-size canvases in word-level models lead to variations in scale and misaligned baselines due to the presence of ascenders and descenders. This makes naïvely concatenating word images unsuitable for producing longer text. To address this, some methods have been explicitly trained on line or paragraph-level data~\cite{davis2020text,kang2021content,pippi2025zero,mayr2024zero}. For length-constrained diffusion models,~\cite{nikolaidou2024diffusionpen} proposed to stitch and blend shorter generations into longer sequences. In this work, we exploit an autoregressive framework without constraints on the output size. 

Following the success of next-token prediction in natural text generation~\cite{vaswani2017attention,radford2019language}, autoregressive image generation has initially been explored by predicting discrete image tokens~\cite{van2017neural,razavi2019generating,esser2021taming, ramesh2021zero,sun2024autoregressive,yu2022scaling}. However, discrete tokenization of images limits generation quality due to compression and optimization problems~\cite{van2017neural,chang2022maskgit,huh2023straightening,mentzer2023finite}, and therefore many recent approaches are evolving towards the prediction of continuous latent vectors~\cite{tschannen2025givt,li2024autoregressive,Infinity,fan2025fluid,team2025nextstep,fan2025unified,zhou2024transfusion}. In line with this progress,~\cite{pippi2025zero} introduced an autoregressive approach that predicts the continous latents of a custom-trained $\beta$-VAE~\cite{higgins2017beta}. To automatically stop the generation at the end of the sentence, the model learns a padding token in the VAE's latent space, stopping generation during inference after 10 consecutive padding tokens. However, this approach is not fully stable as these padding tokens are similar to the tokens representing spaces, and also requires the model to run for more iterations than necessary. In our approach, we introduce a stop token to explicitly stop the generation at the end of the sentence.

Since its introduction~\cite{ho2021classifier}, CFG has been applied to diffusion- and autoregressive-based image generation~\cite{rombach2022high,sun2024autoregressive,quattrini2024merging,quattrini2024alfie,tschannen2025givt} to improve alignment with the target conditioning. CFG is also used in Diffusion HTG models~\cite{nikolaidou2024diffusionpen,mayr2024zero,nikolaidou2025dual,brandenbusch2024semi}. In this work, we apply it to our continuous autoregressive model, adapting it to support the introduced synchronization and stop tokens. This is the first application of this technique in autoregressive HTG.

\section{Eruku Architecture}
\setlength{\belowdisplayskip}{4pt} \setlength{\belowdisplayshortskip}{4pt}
\setlength{\abovedisplayskip}{4pt} \setlength{\abovedisplayshortskip}{4pt}

Our HTG approach takes as input a reference \textit{style image}~($I_s$) and the text content to be rendered, dubbed \textit{generation text}~($T_g$) from here on. Optionally, the model also takes as input the text contained in the style sample, which we call \textit{style text}~($T_s$). Then, it is tasked to generate a text image $I_g$ containing $T_g$ in the same style as $I_s$ (\Cref{fig:pipeline}).

Specifically, we generate $I_g$ by using an autoregressive Transformer Encoder-Decoder model operating in the latent space of a Variational AutoEncoder (VAE), which acts as an image tokenizer. 
The Transformer Encoder, $\mathcal{E}$, takes as input $T_g$, optionally preceded $T_s$. The Transformer Decoder, $\mathcal{D}$, is fed with the $I_s$, tokenized by the VAE Encoder. Then, the model autoregressively outputs embeddings of the VAE latent space and stops generating by emitting a special token. Finally, these latent space tokens are decoded by the VAE Decoder, which outputs the desired text image $I_g$. The details of our pipeline are given below.

\subsection{VAE Image Tokenizer}
To project the style image $I_s$ into a compressed latent space and then convert the embeddings generated by the autoregressive Transformer into the output image $I_g$, we reuse the continuous $\beta$-VAE provided in~\cite{pippi2025zero}, frozen.  
Specifically, given an RGB (${3 \times H \times W}$) text image $I$, the VAE Encoder projects it into a latent space representation. 
Given the VAE's number of channels, $c$, and its downscaling factor, $f$, the resulting latent vector will have shape $c \times h \times w$, where $w=W/f$ and $h=H/f$.  
This is then reshaped into a $w$-long sequence of $h \cdot c$ vectors, $\mathbf{v}=[v_0, ..., v_{w}]$, so that each one encodes a vertical slice of the text image.
The VAE Decoder is tasked to reconstruct a grayscale version of the text image without the background, therefore enforcing the latent space to represent text style rather than background content. The model was trained with a combination of reconstruction loss, KL divergence, Cross-Entropy from a pretrained writing style classification network~\cite{pippi2025zero}, and a CTC loss from a pretrained HTR model~\cite{pippi2025zero}.  

\subsection{Autoregressive Text Encoder}\label{ssec:are}
Eruku autoregressive Transformer Encoder takes as inputs text tokens obtained by tokenizing $T_s$ (if given) and $T_g$ at character-level with the byte-by-byte tokenizer from ByT5~\cite{xue2022byt5}, separated and followed by two special tokens. Specifically, the input to the Encoder is:
\begin{equation*}
\mathbf{t}=[t_{s,1}, ..., t_{s,l_s}, \texttt{<SOG>}, t_{g,1}, ..., t_{g,l_g}, \texttt{<EOG>}].
\end{equation*}
The \textit{textual End of Generation} token \texttt{<EOG>} is a utility token automatically inserted by the tokenizer. The \textit{textual Start of Generation} token \texttt{<SOG>} is an additional special token that we introduce to signal the model that the following tokens are part of the generated image it is tasked to render. The \texttt{<SOG>}, coupled with appropriate training, enables the model to learn to ignore the style text if it is unable to match it to the style image, or if the style text is unavailable. 
The Encoder performs multi-layer self-attention on $\mathbf{t}$ and passes its output to the Autoregressive Decoder.

\subsection{Autoregressive Image Decoder}\label{ssec:ard} 
Eruku autoregressive Transformer Decoder takes as inputs the visual tokens $\mathbf{v}_s$ obtained by tokenizing the style image $I_s$ with the VAE Encoder, linearly projected into embedding vectors with size compatible with the Transformer dimension ($\mathcal{D}_{dim}$), thus obtaining the sequence $\mathbf{e}_s$. Moreover, we prepend and append two additional learnable embeddings: one for the Start of sequence token, $e_{\texttt{SOS}}$, and one for the visual Start of generation token, $e_{\texttt{SOG}}$. Therefore, the input to the Decoder is $\mathbf{e}=[e_\texttt{SOS}, e_{s,1}, ..., e_{s, w_s}, e_\texttt{SOG}]$.

The Decoder performs multi-layer self-attention on $\mathbf{e}$ and cross-attention between $\mathbf{e}$ and the output of the Transformer Encoder, and iteratively generates a sequence of embeddings $\hat{e}_{g,i} \in \mathbf{\hat{e}}$. 
At each generation step, the current embedding $\hat{e}_{g,i}$ is linearly projected into two separate vectors: $\hat{s}_i\in \mathbb{R}^{3}$ and $\hat{v}_{g,i}\in \mathbb{R}^{c\cdot h}$. 
The value of $\hat{s}_i$ belongs to a dictionary of three elements, $\{ $\texttt{<SOG>}$, $\texttt{<IMG>}$, $\texttt{<EOG>}$\}$, within a next-token prediction scheme. In particular: 
\begin{itemize}
    \item Visual Start of Generation token $\texttt{<SOG>}$, meaning that the model suggests to insert the start of generation token at that point in the sequence. This behavior is mainly useful in training, but it also enables the model to recover cases in which the $T_s$ is incorrect. 
    In this case, we append $e_{\texttt{SOG}}$ to the generated sequence of embeddings and run another generation step.
    \item Visual token $\texttt{<IMG>}$, meaning that the model is continuing to generate image tokens in $\hat{v}_i$. In this case, we append its linear projection $\hat{e}_{g,i}$ to the generated sequence and run another generation step.
    \item Visual End of Generation token $\texttt{<EOG>}$ meaning that the model suggests that the entire generation text $T_g$ has been rendered in the output image. In this case, we stop the autoregressive generation process.
\end{itemize}
Note that the visual $\texttt{<SOG>}$ and $\texttt{<EOG>}$ tokens also act as a synchronization signal alongside the corresponding textual $\texttt{<SOG>}$ and $\texttt{<EOG>}$ tokens in the Autoregressive Encoder (see~\Cref{ssec:ard}). 
In fact, as detailed in~\Cref{sec:training}, during training Eruku consistently observes paired textual/visual $\texttt{<SOG>}$ and textual/visual $\texttt{<EOG>}$ tokens. As a consequence, it implicitly learns an alignment between the boundaries of corresponding textual and visual segments. At inference time, the model receives the visual $\texttt{<SOG>}$ already aligned with the textual $\texttt{<SOG>}$, and is encourages to produce the appropriate number of visual tokens to ensure that the textual $\texttt{<EOG>}$ provided to the encoder corresponds meaningfully to the visual $\texttt{<EOG>}$ generated by the decoder.
In the rest of the paper, where it is clear from the context, we use $\texttt{<SOG>}$ and $\texttt{<EOG>}$ for both the textual and visual tokens for simplicity of notation. 

The generation process continues until either a visual $\texttt{<EOG>}$ token is predicted. 
At the end of the process, the sequence of $\hat{v}_{g,i}$'s, $\mathbf{v}_g$, is passed to the VAE Decoder to obtain the final generated image $I_g$.

\tit{Text Classifier-Free Guidance} 
In order to improve Eruku's ability to correctly render the desired text $T_g$ within the output image, we use an inference-time strategy based on Classifier-Free Guidance (CFG). 
CFG was developed in the context of conditional Diffusion Models~\cite{ho2021classifier} to increase prompt-image alignment by sharpening the sampling distribution towards the conditioning. 
To this end, the model is tasked to generate with the conditioning signal (conditional generation) and with a null condition $\emptyset$ (unconditional generation), and the results are combined, scaled with a parameter $\gamma$ to regulate the intensity of this operation.
Recently, this technique has been successfully applied in Autoregressive image generation models~\cite{li2024autoregressive} and in Autoregressive image editing models~\cite{mu2025editar,chen2025context}. 

Recall that our proposed Eruku is conditioned on three inputs: the sequence of embeddings representing the style image, $\mathbf{e}_s$, the sequence of style text tokens $\mathbf{t}_s$, and that of the generation text tokens, $\mathbf{t}_g$. The last two are gathered in a single sequence $\mathbf{t}$. 
To enforce content adherence in the generated image, we apply the CFG formula to the $\hat{e}_{g,i}$'s, but we keep the style image conditioning in the unconditional generation to retain style consistency also in the unconditional branch, \ie
\begin{equation*}
    \begin{aligned}
        p(\hat{e}_{g,i} | \hat{e}_{g,{<i}}&, \mathbf{e}_s, \mathbf{t}) = \\ 
        & p(\hat{e}_{g,i} | \hat{e}_{g,{<i}}, \mathbf{e}_s, \emptyset)~+ \\
        & \gamma \cdot (p(\hat{e}_{g,i} | \hat{e}_{g,{<i}}, \mathbf{e}_s, \mathbf{t}) - (p(\hat{e}_{g,i} | \hat{e}_{g,{<i}}, \mathbf{e}_s, \emptyset) ).
    \end{aligned}
\end{equation*}

\section{Eruku Training}\label{sec:training}

Note that the training samples for our model consist of tuples containing the style text $T_g$, the generation text $T_g$, the style image $I_s$, and the target text image $I_g$. We build a dataset by synthesizing such samples as detailed in~\Cref{ssec:dataset}, and we train our model in two phases, as depicted in~\Cref{fig:pipeline} and described in the following.

\subsection{Training Strategy}
At training time, the input to the Transformer Encoder is the same as what is given at inference, \ie~a sequence of textual tokens $\textbf{t}$, computed from $T_s$ and $T_g$ as explained in \Cref{ssec:are}. 
For the training input to the Transformer Decoder, $I_s$ and $I_g$ are converted to sequences of vectors in the VAE's latent space, \ie~$\textbf{v}_s$ and $\textbf{v}_g$, and projected into sequences of image embeddings $\textbf{e}_s$ and $\textbf{e}_g$. By adding the embeddings of the special visual tokens, we obtain 
\begin{equation*}
    \mathbf{e}=[e_\texttt{SOS}, e_{s,1}, ..., e_{s, w_s}, e_\texttt{SOG}, e_{g,1}, ..., e_{g,w_g}, e_\texttt{EOG}].
\end{equation*}
Then, the model is tasked to replicate the entire sequence of embeddings $\mathbf{e}$, excluding $e_\texttt{SOS}$, \ie~to iteratively output the sequence of vectors
\begin{equation*}
    \mathbf{\hat{e}}=[\hat{e}_{s,1}, ..., \hat{e}_{s, w_s}, \hat{e}_\texttt{SOG}, \hat{e}_{g,1}, ..., \hat{e}_{g,w_g}, \hat{e}_\texttt{EOG}].
\end{equation*}
For the generation during training we apply a teacher-forcing strategy. 

From each $\hat{e}_{i} \in \mathbf{\hat{e}}$, we obtain the corresponding $\hat{s}_{i}$ and $\hat{v}_{i}$ vectors. Then, we compute a Cross-Entropy (CE) loss $\mathcal{L}_{CE}$ on the $\hat{s}_{i}$'s and a Mean Square Error (MSE) loss $\mathcal{L}_{MSE}$ on the  $\hat{v}_{i}$'s. 
Specifically, the $\mathcal{L}_{CE}$ is computed with respect to the reference values in the ground truth sequence given by:
\begin{equation*}
\mathbf{s}=[\texttt{<IMG>}_{\times w_s},\texttt{<SOG>},\texttt{<IMG>}_{\times w_g},\texttt{<EOG>}]. 
\end{equation*} 
This sequence is also used to select the $\hat{v}_{i}$'s corresponding to the style image, $\hat{v}_{s,i}$, and to the desired text image, $\hat{v}_{g,i}$. Then, the $\mathcal{L}_{MSE}$ is computed respectively between the $\hat{v}_{s,i}$'s and the corresponding vector in $\mathbf{v}_s$, and between the $\hat{v}_{g,i}$'s and the corresponding vector in $\mathbf{v}_g$.

\tit{Training for Classifier-Free-Guidance} 
To enable the text CFG as described in \Cref{ssec:ard}, we need to train the model to generate also without conditioning inputs. In our case, this would mean generating without any textual input but only the conditioning given by the style image $I_g$. 
To this end, with a given probability $p_{uncond}$ during training, we replace all text embeddings in the models' textual input $\mathbf{t}$ with a learnable text unconditional embedding, \texttt{<UNCOND>}. In this way, we obtain the $\emptyset$ conditioning for the unconditional generation. We refer to this setting as text-unconditional generation.

\tit{Second Training Phase}
After pre-training, we fine-tune Eruku to enable it to generate also when the style text is not available.
To this end, with a given probability $p_{drop}$ during training, we do not feed the model with the tokens corresponding to $T_s$, \ie~the input to the Transformer Encoder becomes 
\begin{equation*}
\mathbf{t}=[\texttt{<SOG>}, t_{g,1}, ..., t_{g,l_g}, \texttt{<EOG>}].
\end{equation*}
Moreover, in this phase, we use samples whose length is more varied compared to those used in the first training phase. This allows the model to handle longer sequences, which is overall beneficial also in terms of performance~\cite{pippi2025zero}. Text-unconditional training is also performed during this second phase in order to preserve unconditional generation capabilities, as suggested by \cite{phunyaphibarn2025unconditionalpriorsmatterimproving}.

\subsection{Training Data}\label{ssec:dataset}
Commonly, HTG models are trained on a single dataset, hindering their generalization capabilities on out-of-distribution styles, words, and languages. Therefore, we train our model on a specifically-prepared massive and varied synthetic dataset\footnote{https://hf.co/datasets/blowing-up-groundhogs/font-square-pretrain-20M}. 
To obtain the dataset, we collect over 100k typewritten and calligraphic fonts available online and use them to render text words as greyscale ink over a white background. Then, we apply random geometric transformations and split the resulting image to obtain the style image and the generated image, as in~\cite{pippi2023evaluating}. 
The rendered words are picked from a large corpus of English and random words as the one used in~\cite{pippi2025zero}. Specifically, For the pre-training stage, we generate samples consisting of 2 to 3 words for the style images and 2 to 3 words for the target text image, and obtain ~23M samples. 
For the fine-tuning stage, we synthesise ~10M samples whose style image contains 1 to 8 words and the target text image contains 1 to 32 words. 

\section{Experiments}
\label{sec:experiments}

Following~\cite{vanherle2024vatr++,pippi2025zero}, we re-run previous approaches, using their publicly released weights, under a unified setup to enable a fair comparison with the State-of-the-Art. For all test datasets considered, we maintain a fixed set of reference style images and target texts to guide generation, ensuring consistency across methods.

\tit{Implementation details}
We use the pretrained weights from the Emuru VAE~\cite{pippi2025zero}, which has 4 Encoder layers and 4 Decoder layers, a downscaling factor $f=8$ and one output channel ($c=1$). 
The Autoregressive Transformer architecture or Eruku is the same as T5-Large~\cite{raffel2020exploring}, with $\mathcal{E}_{dim}=\mathcal{D}_{dim}=1024$. 
During training, we pad the target images of samples within the same batch so that they have all the same, inter-batch maximum length. For padding, we use the visual \texttt{<EOG>} to teach the model that, once all the visual tokens for $I_g$ are generated, the \texttt{<EOG>} is the only possible output. 
The first training phase is performed with a batch size of 128 over 65000 iterations, whereas the second phase lasts 5000 iterations with a batch size of 2. In both phases, we use gradient accumulation with a virtual batch size of 256, AdamW as optimizer, with a learning rate of 1e-4, and weight decay 1e-2.

\tit{Evaluation Scores}
To comprehensively assess the performance of our model, we employ multiple scores capturing different aspects of HTG. 
These include the task-specific Handwriting Distance (\textbf{HWD})~\cite{pippi2023hwd}, to capture style fidelity, the Absolute Character Error Rate Difference (\textbf{$\Delta$CER})~\cite{pippi2025zero}, which quantifies readability relative to the reference style, the standard image quality evaluation Fréchet Inception Distance (\textbf{FID})~\cite{heusel2017gans}, and the binarized version of the FID, (\textbf{BFID})~\cite{pippi2025zero}, which focuses on font fidelity disregarding the background color and texture~\cite{quattrini2024binarizing}. 

\tit{Datasets}
Our proposed Eruku is trained only on a large synthetic dataset of images containing English text rendered in calligraphy and typewritten fonts, the same as the one use in~\cite{pippi2025zero}. To evaluate its generalization performance, we apply Eruku directly, \ie~without fine-tuning, on multiple multi-writer datasets. These include the \textbf{IAM}~\cite{marti2002iam} dataset (both word- and line-level), and the line-level \textbf{CVL}~\cite{kleber2013cvl} and \textbf{RIMES}~\cite{augustin2006rimes} datasets. Moreover, we consider the line-level \textbf{Karaoke}~\cite{pippi2025zero} dataset, consisting of song lyrics in English, French, German, and Italian rendered using 100 publicly available fonts\footnote{\url{https://fonts.google.com/}}, encompassing both calligraphy and typewritten styles on a white background.

\tit{Compared Methods}
We compare Eruku against State-of-the-Art HTG methods with publicly released code and pretrained weights. Specifically, we include Convolutional GAN-based models \textbf{HiGAN+}~\cite{gan2022higan+} and \textbf{TS-GAN}~\cite{davis2020text}, Transformer GAN-based methods \textbf{HWT}~\cite{bhunia2021handwriting}, \textbf{VATr}~\cite{pippi2023handwritten}, and \textbf{VATr++}~\cite{vanherle2024vatr++}, as well as diffusion-based approaches \textbf{DiffPen}~\cite{nikolaidou2024diffusionpen} and \textbf{One-DM}~\cite{dai2024one}. Finally, we consider the Autoregressive Transformer-based \textbf{Emuru}~\cite{pippi2025zero}, which is the closest to our approach.

\subsection{Results}
First, we perform ablation analyses of Eruku's main characteristics. To this end, we consider the line-level version of the IAM dataset, since it is the most commonly adopted in HTG literature. 

\tit{Style Text Drop}
We validate the effect of dropping the style text $T_s$ in the Eruku input. In~\Cref{tab:style_drop}, we report the results obtained by applying varying style text drop probability $p_{drop}$ in the second phase of training. For this comparison, we generate both with and without $T_s$ and use $\gamma=1.25$ for the text CFG. Note that the baseline model is the first line in the same~\Cref{tab:style_drop}, \ie~our Eruku architecture trained with style text $T_s$, which is also provided during inference. 
Observing the results, we can see that the baseline model, which was not trained with style text dropout ($p_{drop}=0$), does not work well when the style text is not provided during inference. This is expected, as it has never been trained in this setting. 
As intended, the ability of the model to generate without style text is significantly improved with increasing values of $p_{drop}$. In particular, moving from $p_{drop}=0$ to $p_{drop}=1$ yields significant performance improvements when style text is not provided. An added benefit of this style text dropout strategy is that the model trained with $p_{drop}=0.1$ improves its HWD and FID significantly also when provided with $T_s$, without any decrease in $\Delta$CER. This leads us to believe that this form of style text dropout makes the model more robust to style text that it would otherwise fail to match to the style image. Therefore, the model is able to leverage the provided style text $T_s$ better than to the baseline model.

With increasing values of $p_{drop}$, we can see that the model's performance increases without the $T_s$ input and decreases when $T_s$ is provided. We attribute this to the model losing the ability to match $T_s$ and $I_g$ if no longer provided with sufficient paired samples in the second phase of training. To show this, we also report the results of a model finetuned with $p_{drop}=1$, which learns quite well how to generate without style text input. When we input $T_s$ to this model, its performance drops. 
In light of these results, we find $p_{drop}=0.1$ to be the best tradeoff between non-style text-conditioned generation and style text-conditioned generation and use this value for our final model.

\begin{table}[t]
    \centering
    \setlength{\tabcolsep}{.8em}
    \resizebox{\columnwidth}{!}{%
    \begin{tabular}{c c cccc}
    \toprule
    $p_{drop}$
    & $T_s$
    & \textbf{HWD$\downarrow$}
    & \textbf{$\Delta$CER$\downarrow$}
    & \textbf{FID$\downarrow$}
    & \textbf{BFID$\downarrow$} \\
    \midrule
       \multirow{2}{*}{0.0} & \checkmark & 1.81 & \textbf{0.48} & 14.20 & 3.88 \\
      & \xmark & 3.03 & 1.02 & 86.97 & 77.46 \\
    \midrule 
      \multirow{2}{*}{0.1} & \checkmark & 1.75 & \textbf{0.48} & \textbf{13.49} & \textbf{4.45} \\
     & \xmark & 2.15 & 0.47 & 16.71 & 8.10 \\
    \midrule 
      \multirow{2}{*}{0.3} & \checkmark & 1.79 & 0.50 & 16.67 & 4.62 \\
     & \xmark & 2.17 & 0.46 & 18.73 & 5.00 \\
    \midrule 
      \multirow{2}{*}{0.6} & \checkmark & \textbf{1.74} & 0.51 & 15.34 & 5.47 \\
     & \xmark & 2.05 & 0.46 & 16.91 & 6.11 \\
    \midrule 
      \multirow{2}{*}{1.0} & \checkmark & 2.02 & 0.75 & 21.41 & 15.63 \\
     & \xmark & 2.03 & 0.38 & 15.97 & 4.25 \\
    \bottomrule
    \end{tabular}
}\caption{Effect of the style text drop probability applied in training, $p_{drop}$, on the performance of Eruku on IAM Lines, both when the style text $T_s$ is given or not at inference time. For reference, we also report the result of Eruku fed with $T_s$ obtained by running TrOCR on the style image $I_s$ (dubbed $T_s^*$).\vspace{-0.5cm}}
\label{tab:style_drop}
\end{table}

\tit{Text-CFG} 
To validate the effect of the proposed text-CFG on the model's performance, we consider its effect at different values of the scale $\gamma$ at inference time. As a reference, we also consider the performance obtained by a variant of our approach not trained to perform text-unconditional generation, which is therefore ran without CFG at inference time. The results of this analysis are reported in~\Cref{tab:cfg}. 
We can observe that increasing the CFG scale directly impacts the $\Delta$CER, and therefore the correctness and readability of the text within the output image. For values of $\gamma$ up to 1.25, this improvement does not hinder the style adherence measured by HWD. When increasing beyond that value, there are diminishing advantages in terms of $\Delta$CER, and worsening HWD values, indicating that style adherence decreases as $\gamma$ increases beyond 1.25. For this reason, we select $\gamma=1.25$ as the default CFG scale value for Eruku. Nonetheless, a user could change this CFG scale value to obtain a different style-text adherence trade-off.

\begin{figure}
    \centering
    \includegraphics[width=\linewidth]{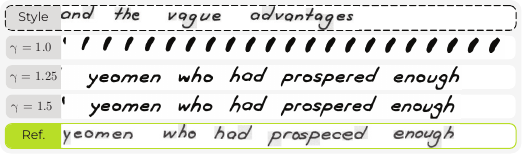}
    \caption{Qualitative analysis of the CFG effect on generation at varying $\gamma$'s. We generate by giving as $I_s$ the top image in the figure (Style) and as $T_s$ the text contained in the bottom image (Ref.).
    }
    \label{fig:cfg}
\end{figure}

Note that an alternative, popular way in the HTG literature to improve the readability of the text output is to fine-tune the models using an auxiliary HTR network~\cite{pippi2023handwritten, vanherle2024vatr++, nikolaidou2024diffusionpen}. As an ablation, we also try the same approach on Eruku by introducing, in the second stage of training, supervision from an OCR model trained~\cite{yousef2020origaminet} on the same synthetic data used for Eruku in that phase of training. 
The results obtained are reported in~\Cref{tab:cfg} by adding the suffix ``+ OCR''. We observe that OCR fine-tuning proves to be less style-preserving than CFG at obtaining the same $\Delta$CER values, as reflected in the HWD scores. Therefore, we do not perform OCR fine-tuning for the final Eruku model, which, as mentioned in \Cref{sec:training}, is trained without any need for auxiliary networks. To further isolate the effect of CFG training, we also train Eruku's variants without conditioning dropout, and then fine-tune using the OCR model~\cite{yousef2020origaminet}. The results, reported in~\Cref{tab:cfg}, show that these variants are worse in both style preservation and text correctness. Finally, we qualitatively show the effect of $\gamma$ in~\Cref{fig:cfg}.

\begin{table}[t]
    \centering
    \setlength{\tabcolsep}{.3em}
    \resizebox{\columnwidth}{!}{%
    \begin{tabular}{l cccc}
    \toprule
    & \textbf{HWD$\downarrow$}
    & \textbf{$\Delta$CER$\downarrow$}
    & \textbf{FID$\downarrow$}
    & \textbf{BFID$\downarrow$} \\
    \midrule
    \textbf{Eruku ($\gamma=1$)}     & 1.83 & 0.18 & 20.12 & 12.52 \\
    \textbf{Eruku ($\gamma=1.125$)} & 1.73 & 0.10 & 17.44 & 7.71 \\
    \textbf{Eruku ($\gamma=1.25$)}  & \textbf{1.70} & 0.06 & \textbf{16.40} & 4.88 \\
    \textbf{Eruku ($\gamma=1.375$)} & 1.73 & \textbf{0.04} & 16.76 & \textbf{4.01} \\
    \textbf{Eruku ($\gamma=1.5$)}   & 1.80 & \textbf{0.04} & 17.51 & 4.07 \\
    \midrule
    \textbf{Eruku ($\gamma=1$)  + OCR}     & 1.82 & 0.14 & 16.02 & 7.99 \\
    \textbf{Eruku ($\gamma=1.125$) + OCR} & \textbf{1.75} & 0.08 & \textbf{15.27} & 4.99 \\
    \textbf{Eruku ($\gamma=1.25$) + OCR}  & 1.78 & 0.05 & 16.45 & \textbf{4.86} \\
    \textbf{Eruku ($\gamma=1.375$) + OCR} & 1.84 & 0.04 & 18.04 & 5.18 \\
    \textbf{Eruku ($\gamma=1.5$) + OCR}   & 1.91 & \textbf{0.03} & 19.66 & 6.02 \\
    \midrule    
    \textbf{Eruku*}                 & 1.83 & 0.21 & 21.10 & 17.11 \\
    \textbf{Eruku* + OCR}           & \textbf{1.82} & \textbf{0.18} & \textbf{18.48} & \textbf{11.25} \\
    \bottomrule
    \end{tabular}
}\caption{Effect of the text CFG scale on the performance of Eruku.  $\gamma=1$ means that the text CFG is not performed. We also report the results of a variant not trained to support the text CFG (dubbed Eruku*) and variants trained with the additional supervision of an auxiliary OCR network (noted `+ OCR').}\vspace{-.3cm}
\label{tab:cfg}
\end{table}

\begin{figure*}
    \centering
    \includegraphics[width=\linewidth]{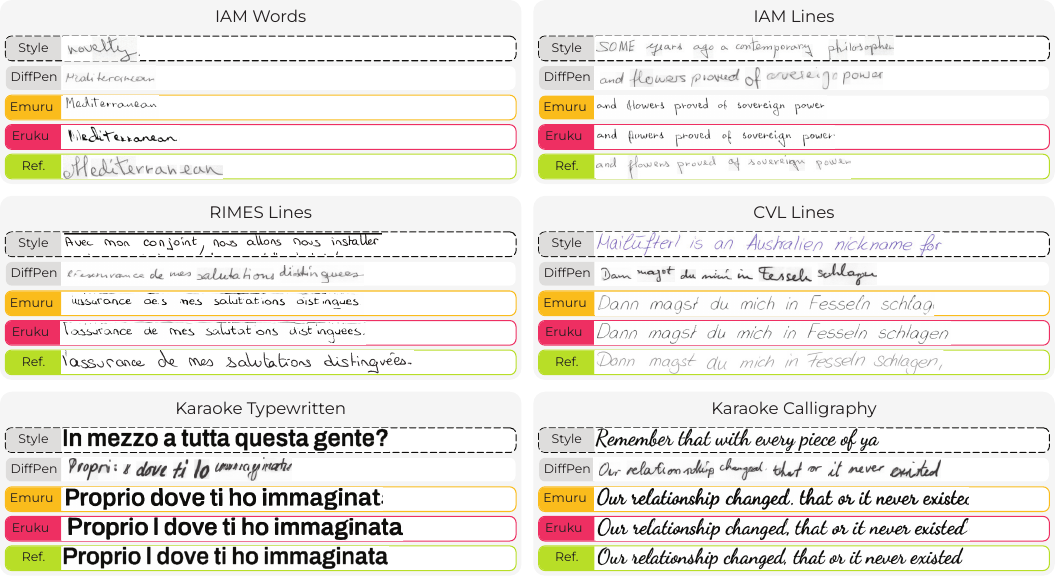}
    \caption{Qualitative results between our proposed Eruku, and the State-of-the-Art Emuru and DiffPen models on the considered datasets. We task the models to generate a replica of the reported reference image (Ref.) by giving them as input the text contained in Ref. and the reported style image (Style).}
    \label{fig:qualitatives}
\end{figure*}

\tit{Comparison with the State-of-the-Art}
In~\Cref{tab:iam,tab:cvl_rimes,tab:karaoke} and~\Cref{fig:qualitatives}, we report a quantitative and qualitative comparison between Eruku and other State-of-the-Art HTG approaches. The comparison is performed both on the dataset that most of the competitors have seen in training (IAM) and on unseen datasets. Recall that, instead, for Emuru and our Eruku approach, all the datasets are unseen. Our model exhibits strong generalization capabilities, maintaining solid performance on all line-level datasets. This is underscored by the fact that it is the best-performing model in terms of HWD on all datasets except for IAM Words. The lower performance on this word-level dataset could be attributed to the particular emphasis given during training to long-context generation in the second stage. A potential mitigation that could be implemented in future approaches is to supplement the dataset used for long-context training with a larger sample of short images. consisting of one or very few words.

\begin{table*}[t]
\centering
\setlength{\tabcolsep}{.25em}
\begin{minipage}[t]{0.32\textwidth}
    \centering
    \resizebox{\columnwidth}{!}{%
        \begin{tabular}{l cccc}
    \toprule
    & \multicolumn{4}{c}{\textbf{IAM Words}}\\ 
    \cmidrule{2-5}
    & \textbf{HWD$\downarrow$}
    & \textbf{$\Delta$CER$\downarrow$}
    & \textbf{FID$\downarrow$}
    & \textbf{BFID$\downarrow$} \\
    \midrule
    \textbf{TS-GAN}          & 4.22 & 0.28 & 129.57 & 86.45 \\
    \textbf{HiGAN+}          & 3.12 & 0.20 & ~50.19 & 21.92 \\
    \textbf{HWT}             & 2.01 & 0.15 & ~27.83 & 15.09 \\
    \textbf{VATr}            & 2.19 & \textbf{0.00} & ~30.26 & 15.81 \\
    \textbf{VATr++}          & 2.54 & 0.07 & ~31.91 & 17.15 \\
    \textbf{One-DM}          & 2.28 & 0.10 & ~27.54 & 10.73 \\
    \textbf{DiffPen}         & \textbf{1.78} & 0.06 & ~\textbf{15.54} & ~\textbf{6.06} \\
    \textbf{Emuru}           & 3.03 & 0.19 & ~63.61 & 37.73 \\
    \textbf{Eruku}           & 3.23 & 0.77 & ~79.66 & 63.31 \\
    \midrule
        & \multicolumn{4}{c}{\textbf{IAM Lines}}\\ 
    \cmidrule{2-5}
    & \textbf{HWD$\downarrow$}
    & \textbf{$\Delta$CER$\downarrow$}
    & \textbf{FID$\downarrow$}
    & \textbf{BFID$\downarrow$}\\
    \midrule
    \textbf{TS-GAN}          & 3.21 & 0.02 & 44.17 & 19.45 \\
    \textbf{HiGAN+}          & 3.25 & \textbf{0.00} & 74.41 & 34.18 \\
    \textbf{HWT}             & 2.97 & 0.33 & 44.72 & 30.26 \\
    \textbf{VATr}            & 2.37 & 0.02 & 35.32 & 27.97 \\
    \textbf{VATr++}          & 2.38 & 0.03 & 34.00 & 21.67 \\
    \textbf{One-DM}          & 2.83 & 0.13 & 43.89 & 21.54 \\
    \textbf{DiffPen}         & 2.13 & 0.03 & \textbf{12.89} & ~6.87 \\
    \textbf{Emuru}           & {1.87} & 0.14 & 13.89 & ~6.19 \\ 
    \textbf{Eruku}           & \textbf{1.70} & 0.06 & 16.40 & ~\textbf{4.88} \\
    \bottomrule
    \end{tabular}
    }
    \caption{Comparison on the word-level and line-level IAM datasets. Note that Eruku and Emuru have not been trained on IAM.}
    \label{tab:iam}
\end{minipage}
\hfill
\begin{minipage}[t]{0.32\textwidth}
    \centering
    \resizebox{\columnwidth}{!}{%
        \begin{tabular}{l cccc}
    \toprule
    & \multicolumn{4}{c}{\textbf{CVL Lines}}\\ 
    \cmidrule{2-5}
    & \textbf{HWD$\downarrow$}
    & \textbf{$\Delta$CER$\downarrow$}
    & \textbf{FID$\downarrow$}
    & \textbf{BFID$\downarrow$} \\
    \midrule
    \textbf{TS-GAN}          & 3.07 & 0.13 & 42.12 & 31.97 \\
    \textbf{HiGAN+}          & 3.07 & 0.12 & 78.44 & 39.47 \\
    \textbf{HWT}             & 2.59 & 0.38 & 31.22 & 16.73 \\
    \textbf{VATr}            & 2.36 & 0.06 & 34.40 & 24.64 \\
    \textbf{VATr++}          & 2.18 & 0.12 & 35.53 & 19.87 \\
    \textbf{One-DM}          & 2.66 & 0.06 & 60.45 & 26.58 \\
    \textbf{DiffPen}         & 2.99 & \textbf{0.01} & 40.40 & 17.50 \\
    \textbf{Emuru}           & 1.82 & 0.13 & 14.39 & 10.77 \\
    \textbf{Eruku}           & \textbf{1.72} & 0.04 & \textbf{12.32} & ~\textbf{6.62} \\
    \midrule
        & \multicolumn{4}{c}{\textbf{RIMES Lines}}\\ 
    \cmidrule{2-5}
    & \textbf{HWD$\downarrow$}
    & \textbf{$\Delta$CER$\downarrow$}
    & \textbf{FID$\downarrow$}
    & \textbf{BFID$\downarrow$}\\
    \midrule
    \textbf{TS-GAN}          & 3.26 & 0.12 & 109.04 & 36.39 \\
    \textbf{HiGAN+}          & 3.39 & 0.14 & 160.57 & 47.38 \\
    \textbf{HWT}             & 3.36 & 0.45 & 118.21 & 35.26 \\
    \textbf{VATr}            & 3.09 & 0.07 & 113.76 & 30.21 \\
    \textbf{VATr++}          & 2.83 & 0.10 & 110.04 & 35.61 \\
    \textbf{One-DM}          & 3.36 & 0.20 & 121.18 & 36.07 \\
    \textbf{DiffPen}         & 2.58 & \textbf{0.04} & ~89.79 & 18.25 \\
    \textbf{Emuru}           & 2.18 & 0.25 & ~\textbf{26.93} & 13.26 \\
    \textbf{Eruku}           & \textbf{1.81} & 0.11 & ~27.51 & \textbf{10.15} \\
    \bottomrule
    \end{tabular}
    }
    \caption{Comparison on the CVL and RIMES datasets. Note that none of the approaches has been trained on these datasets.}
    \label{tab:cvl_rimes}
\end{minipage}
\hfill
\begin{minipage}[t]{0.32\textwidth}
    \centering
    \resizebox{\columnwidth}{!}{%
        \begin{tabular}{l cccc}
    \toprule
    & \multicolumn{4}{c}{\textbf{Karaoke Calligraphy}}\\ 
    \cmidrule{2-5}
    & \textbf{HWD$\downarrow$}
    & \textbf{$\Delta$CER$\downarrow$}
    & \textbf{FID$\downarrow$}
    & \textbf{BFID$\downarrow$} \\
    \midrule
    \textbf{TS-GAN}          & 4.59 & 0.23 & 60.30 & 12.68 \\
    \textbf{HiGAN+}          & 4.90 & 0.08 & 125.75 & 69.41 \\
    \textbf{HWT}             & 4.50 & 0.32 & 62.69 & 43.03 \\
    \textbf{VATr}            & 3.89 & 0.05 & 72.22 & 47.66 \\
    \textbf{VATr++}          & 3.96 & \textbf{0.01} & 67.16 & 46.53 \\
    \textbf{One-DM}          & 4.31 & 0.04 & 59.73 & 38.30 \\
    \textbf{DiffPen}         & 4.18 & 0.16 & 34.19 & 25.78 \\
    \textbf{Emuru}           & 2.24 & 0.13 & 13.87 & ~7.99 \\
    \textbf{Eruku}           & \textbf{2.04} & 0.13 & \textbf{12.39} & ~\textbf{7.30} \\
    \midrule
        & \multicolumn{4}{c}{\textbf{Karaoke Typewritten}}\\ 
    \cmidrule{2-5}
    & \textbf{HWD$\downarrow$}
    & \textbf{$\Delta$CER$\downarrow$}
    & \textbf{FID$\downarrow$}
    & \textbf{BFID$\downarrow$}\\
    \midrule
    \textbf{TS-GAN}          & 4.70 & 0.32 & 141.41 & 75.78 \\
    \textbf{HiGAN+}          & 5.19 & 0.07 & 135.34 & 63.39 \\
    \textbf{HWT}             & 4.57 & 0.37 & 72.78 & 37.40 \\
    \textbf{VATr}            & 4.14 & 0.05 & 80.38 & 41.02 \\
    \textbf{VATr++}          & 4.15 & \textbf{0.01} & 76.03 & 41.69 \\
    \textbf{One-DM}          & 4.80 & 0.05 & 70.75 & 44.06 \\
    \textbf{DiffPen}         & 4.71 & 0.14 & 78.07 & 61.16 \\
    \textbf{Emuru}           & 1.28 & 0.11 & ~\textbf{9.85} & ~\textbf{4.33} \\
    \textbf{Eruku}           & \textbf{1.21} & 0.11 & 10.29 & ~5.07 \\
    \bottomrule
    \end{tabular}
    }
    \caption{Comparison on the Karaoke dataset. Note that none of the approaches has been trained on these datasets.}
    \label{tab:karaoke}
\end{minipage}
\end{table*}

\section{Conclusion}
\label{sec:conclusion}

In this paper, we have tackled the limitations of the current State-of-the-Art Autoregressive HTG approach, namely the dependency on accurate style image transcriptions, the inefficient and error-prone stopping mechanism, and the poor adherence to the target text to render. To this end, we have introduced Eruku, an Autoregressive Transformer Encoder-Decoder that incorporates special visual and textual tokens and a novel CFG-inspired mechanism acting on the text inputs. Through extensive experiments on a variety of handwritten and typewritten datasets, we validate the effectiveness of our approach in operating even with missing or noisy style image transcriptions and in generating images whose content closely adheres to the desired target text, while maintaining generalization capabilities and output length flexibility. 

\section*{Acknowledgement}
This paper is based upon work supported by the GCP Credit Award, the Google Cloud Research Credits program with the award GCP19980904, and the FARD2025 project (CUP E93C25000370005).
We acknowledge EuroHPC Joint Undertaking and ISCRA for awarding us access to LUMI at CSC, Finland, LEONARDO at CINECA, Italy, and MareNostrum5 at BSC, Spain. 


{
    \small
    \bibliographystyle{ieeenat_fullname}
    \bibliography{main}

@string{jmlr      = {Journal of Machine Learning Research}}

@string{nips      = {Advances in Neural Information Processing Systems}}

@string{iccv      = {Proceedings of the IEEE/CVF International Conference on Computer Vision}}

@string{iccvw     = {Proceedings of the IEEE/CVF International Conference on Computer Vision Workshops}}

@string{eccv      = {Proceedings of the European Conference on Computer Vision}}

@string{eccvw     = {Proceedings of the European Conference on Computer Vision Workshops}}

@string{cvpr      = {Proceedings of the IEEE/CVF Conference on Computer Vision and Pattern Recognition}}

@string{iclr      = {Proceedings of the International Conference on Learning Representations}}

@string{icml      = {Proceedings of the International Conference on Machine Learning}}

@string{bmvc      = {Proceedings of the British Machine Vision Conference}}

@string{aaai      = {Proceedings of the AAAI Conference on Artificial Intelligence}}

@string{icdar     = {Proceedings of the International Conference on Document Analysis and Recognition}}

@string{iwfhr     = {Proceedings of the International Workshop on Frontiers in Handwriting Recognition}}

@string{chi       = {Proceedings of the CHI Conference on Human Factors in Computing Systems}}

@string{ieeetpami = {IEEE Transactions on Pattern Analysis and Machine Intelligence}}

@string{ieeetnnls  = {IEEE Transactions on Neural Networks and Learning Systems}}

@string{pr        = {Pattern Recognition}}

@string{prl       = {Pattern Recognition Letters}}

@string{ijdar     = {International Journal on Document Analysis and Recognition}}

@string{acmtog    = {ACM Transactions on Graphics}}

@string{iconip = {International Conference on Neural Information Processing}}

@string{tmlr = {Transactions on Machine Learning Research}}

@string{tacl    = {Transactions of the Association for Computational Linguistics}}

@string{nips     = {NeurIPS}}

@string{iccv     = {ICCV}}

@string{cvpr     = {CVPR}}

@string{iclr     = {ICLR}}

@string{bmvc     = {BMVC}}

@string{eccv     = {ECCV}}

@string{icml     = {ICML}}

@string{aaai     = {AAAI}}

@string{ijdar    = {IJDAR}}

@string{icdar    = {ICDAR}}

@string{iwfhr    = {IWFHR}}

@string{chi       = {CHI}}

@string{iconip = {ICONIP}}

@string{tmlr = {TMLR}}

@string{tacl = {TACL}}

@STRING{ieeetpami  = {IEEE Trans. PAMI}}

@string{ieeetnnls  = {IEEE Trans. Neural Netw. Learn. Syst.}}

@string{ijdar      = {Int. J. Doc. Anal. Recognit.}}

@string{acmtog    = {{ACM Trans. Graphics}}}

@string{prl       = {Pattern Recognit. Lett.}}

@string{pr       = {Pattern Recognit.}}

@string{tacl = {Trans. Assoc. Comput. Linguist.}}

@inproceedings{zhu2023conditional,
  title={{Conditional Text Image Generation with Diffusion Models}},
  author={Zhu, Yuanzhi and Li, Zhaohai and Wang, Tianwei and He, Mengchao and Yao, Cong},
  booktitle=cvpr,
  year={2023}
}

@inproceedings{nikolaidou2023wordstylist,
  title={{WordStylist: Styled Verbatim Handwritten Text Generation with Latent Diffusion Models}},
  author={Nikolaidou, Konstantina and Retsinas, George and Christlein, Vincent and Seuret, Mathias and Sfikas, Giorgos and Smith, Elisa Barney and Mokayed, Hamam and Liwicki, Marcus},
  booktitle=icdar,
  year={2023}
}

@inproceedings{pippi2023handwritten,
  title={{Handwritten Text Generation from Visual Archetypes}},
  author={Pippi, Vittorio and Cascianelli, Silvia and Cucchiara, Rita},
  booktitle=cvpr,
  year={2023}
}

@inproceedings{pippi2023hwd,
  title={{HWD: A Novel Evaluation Score for Styled Handwritten Text Generation}},
  author={Pippi, Vittorio and Quattrini, Fabio and Cascianelli, Silvia and Cucchiara, Rita},
  booktitle=bmvc,
  year={2023}
}

@inproceedings{pippi2023choose,
  title={How to choose pretrained handwriting recognition models for single writer fine-tuning},
  author={Pippi, Vittorio and Cascianelli, Silvia and Kermorvant, Christopher and Cucchiara, Rita},
  booktitle=icdar,
  year={2023}
}

@inproceedings{kleber2013cvl,
  title={{CVL-DataBase: An Off-Line Database for Writer Retrieval, Writer Identification and Word Spotting}},
  author={Kleber, Florian and Fiel, Stefan and Diem, Markus and Sablatnig, Robert},
  booktitle=icdar,
  year={2013}
}

@inproceedings{heusel2017gans,
  title={{GANs Trained by a Two Time-Scale Update Rule Converge to a Local Nash Equilibrium}},
  author={Heusel, Martin and Ramsauer, Hubert and Unterthiner, Thomas and Nessler, Bernhard and Hochreiter, Sepp},
  booktitle=nips,
  year={2017}
}

@article{cascianelli2022boosting,
  title={{Boosting Modern and Historical Handwritten Text Recognition with Deformable Convolutions}},
  author={Cascianelli, Silvia and Cornia, Marcella and Baraldi, Lorenzo and Cucchiara, Rita and others},
  journal=ijdar,
  pages={1--15},
  year={2022}
}

@inproceedings{aksan2018deepwriting,
  title={{DeepWriting: Making digital ink editable via deep generative modeling}},
  author={Aksan, Emre and Pece, Fabrizio and Hilliges, Otmar},
  booktitle=chi,
  year={2018}
}

@article{graves2013generating,
  title={{Generating Sequences with Recurrent Neural Networks}},
  author={Graves, Alex},
  journal={arXiv preprint arXiv:1308.0850},
  year={2013}
}

@article{haines2016my,
  title={{My Text in Your Handwriting}},
  author={Haines, TSF and Mac Aodha, O and Brostow, GJ},
  journal=acmtog,
  volume={35},
  number={3},
  year={2016},
  publisher={ACM}
}

@inproceedings{aksan2018stcn,
  title={{STCN: Stochastic Temporal Convolutional Networks}},
  author={Aksan, Emre and Hilliges, Otmar},
  booktitle=iclr,
  year={2018}
}

@inproceedings{alonso2019adversarial,
  title={{Adversarial Generation of Handwritten Text Images Conditioned on Sequences}},
  author={Alonso, Eloi and Moysset, Bastien and Messina, Ronaldo},
  booktitle=icdar,
  year={2019}
}

@article{gan2022higan+,
  title={{HiGAN+: Handwriting Imitation GAN with Disentangled Representations}},
  author={Gan, Ji and Wang, Weiqiang and Leng, Jiaxu and Gao, Xinbo},
  journal=acmtog,
  pages={1--17},
  year={2022}
}

@article{kang2021content,
  title={Content and Style Aware Generation of Text-line Images for Handwriting Recognition},
  author={Kang, Lei and Riba, Pau and Rusinol, Marcal and Fornes, Alicia and Villegas, Mauricio},
  journal=ieeetpami,
  pages={1--1},
  year={2021}
}

@article{krishnan2023textstylebrush,
  title={{TextStyleBrush: Transfer of Text Aesthetics from a Single Example}},
  author={Krishnan, Praveen and Kovvuri, Rama and Pang, Guan and Vassilev, Boris and Hassner, Tal},
  journal=ieeetpami,
  year={2023}
}

@inproceedings{bhunia2021handwriting,
  title={{Handwriting Transformers}},
  author={Bhunia, Ankan Kumar and Khan, Salman and Cholakkal, Hisham and Anwer, Rao Muhammad and Khan, Fahad Shahbaz and Shah, Mubarak},
  booktitle=iccv,
  year={2021}
}

@article{luo2022slogan,
  title={{SLOGAN: Handwriting Style Synthesis for Arbitrary-Length and Out-of-Vocabulary Text}},
  author={Luo, Canjie and Zhu, Yuanzhi and Jin, Lianwen and Li, Zhe and Peng, Dezhi},
  journal=ieeetnnls,
  year={2022}
}

@inproceedings{davis2020text,
  title={{Text and Style Conditioned GAN for Generation of Offline Handwriting Lines}},
  author={Davis, Brian and Tensmeyer, Chris and Price, Brian and Wigington, Curtis and Morse, Bryan and Jain, Rajiv},
  booktitle=bmvc,
  year={2020}
}

@inproceedings{kang2020ganwriting,
   title={{GANwriting: Content-Conditioned Generation of Styled Handwritten Word Images}},
  author={Kang, Lei and Riba, Pau and Wang, Yaxing and Rusi{\~n}ol, Mar{\c{c}}al and Forn{\'e}s, Alicia and Villegas, Mauricio},
  booktitle=eccv,
  year={2020}
}

@inproceedings{fogel2020scrabblegan,
  title={{ScrabbleGAN: Semi-Supervised Varying Length Handwritten Text Generation}},
  author={Fogel, Sharon and Averbuch-Elor, Hadar and Cohen, Sarel and Mazor, Shai and Litman, Roee},
  booktitle=cvpr,
  year={2020}
}

@inproceedings{gan2021higan,
  title={{HiGAN: Handwriting Imitation Conditioned on Arbitrary-Length Texts and Disentangled Styles}},
  author={Gan, Ji and Wang, Weiqiang},
  booktitle=aaai,
  year={2021}
}

@article{radford2019language,
  title={{Language models are unsupervised multitask learners}},
  author={Radford, Alec and Wu, Jeffrey and Child, Rewon and Luan, David and Amodei, Dario and Sutskever, Ilya and others},
  journal={OpenAI blog},
  pages={9},
  year={2019}
}

@article{van2017neural,
  title={{Neural discrete representation learning}},
  author={Van Den Oord, Aaron and Vinyals, Oriol and others},
  journal=nips,
  volume={30},
  year={2017}
}

@article{razavi2019generating,
  title={{Generating diverse high-fidelity images with vq-vae-2}},
  author={Razavi, Ali and Van den Oord, Aaron and Vinyals, Oriol},
  journal=nips,
  year={2019}
}

@inproceedings{esser2021taming,
  title={{Taming transformers for high-resolution image synthesis}},
  author={Esser, Patrick and Rombach, Robin and Ommer, Bjorn},
  booktitle=cvpr,
  year={2021}
}

@inproceedings{ramesh2021zero,
  title={{Zero-shot text-to-image generation}},
  author={Ramesh, Aditya and Pavlov, Mikhail and Goh, Gabriel and Gray, Scott and Voss, Chelsea and Radford, Alec and Chen, Mark and Sutskever, Ilya},
  booktitle=icml,
  year={2021},
}

@article{yu2022scaling,
  title={{Scaling autoregressive models for content-rich text-to-image generation}},
  author={Yu, Jiahui and Xu, Yuanzhong and Koh, Jing Yu and Luong, Thang and Baid, Gunjan and Wang, Zirui and Vasudevan, Vijay and Ku, Alexander and Yang, Yinfei and Ayan, Burcu Karagol and others},
  journal=tmlr,
  year={2022}
}

@article{sun2024autoregressive,
  title={{Autoregressive Model Beats Diffusion: Llama for Scalable Image Generation}},
  author={Sun, Peize and Jiang, Yi and Chen, Shoufa and Zhang, Shilong and Peng, Bingyue and Luo, Ping and Yuan, Zehuan},
  journal={arXiv preprint arXiv:2406.06525},
  year={2024}
}

@inproceedings{huh2023straightening,
  title={{Straightening out the straight-through estimator: Overcoming optimization challenges in vector quantized networks}},
  author={Huh, Minyoung and Cheung, Brian and Agrawal, Pulkit and Isola, Phillip},
  booktitle=icml,
  year={2023}
}

@inproceedings{
mentzer2023finite,
title={{Finite Scalar Quantization: {VQ}-{VAE} Made Simple}},
author={Fabian Mentzer and David Minnen and Eirikur Agustsson and Michael Tschannen},
booktitle=iclr,
year={2024}
}

@inproceedings{chang2022maskgit,
  title={{Maskgit: Masked generative image transformer}},
  author={Chang, Huiwen and Zhang, Han and Jiang, Lu and Liu, Ce and Freeman, William T},
  booktitle=cvpr,
  year={2022}
}

@inproceedings{dai2023disentangling,
  title={{Disentangling Writer and Character Styles for Handwriting Generation}},
  author={Dai, Gang and Zhang, Yifan and Wang, Qingfeng and Du, Qing and Yu, Zhuliang and Liu, Zhuoman and Huang, Shuangping},
  booktitle=cvpr,
  year={2023}
}

@article{higgins2017beta,
  title={{beta-vae: Learning basic visual concepts with a constrained variational framework.}},
  author={Higgins, Irina and Matthey, Loic and Pal, Arka and Burgess, Christopher P and Glorot, Xavier and Botvinick, Matthew M and Mohamed, Shakir and Lerchner, Alexander},
  journal=iclr,
  year={2017}
}

@inproceedings{tschannen2025givt,
  title={{Givt: Generative infinite-vocabulary transformers}},
  author={Tschannen, Michael and Eastwood, Cian and Mentzer, Fabian},
  booktitle=eccv,
  year={2025},
}

@article{li2021trocr,
  title={{TrOCR: Transformer-based optical character recognition with pre-trained models}},
  author={Li, Minghao and Lv, Tengchao and Cui, Lei and Lu, Yijuan and Florencio, Dinei and Zhang, Cha and Li, Zhoujun and Wei, Furu},
  journal=aaai,
  year={2023}
}

@inproceedings{yousef2020origaminet,
  title={{OrigamiNet: Weakly-Supervised, Segmentation-Free, One-Step, Full Page Text Recognition by learning to unfold}},
  author={Yousef, Mohamed and Bishop, Tom E},
  booktitle=cvpr,
  year={2020}
}

@article{marti2002iam,
  title={{The IAM-database: an English sentence database for offline handwriting recognition}},
  author={Marti, U-V and Bunke, Horst},
  journal=ijdar,
  pages={39--46},
  year={2002}
}

@inproceedings{vaswani2017attention,
  title={{Attention is all you need}},
  author={Vaswani, Ashish and Shazeer, Noam and Parmar, Niki and Uszkoreit, Jakob and Jones, Llion and Gomez, Aidan N and Kaiser, {\L}ukasz and Polosukhin, Illia},
  booktitle=nips,
  year={2017}
}

@inproceedings{augustin2006rimes,
  title={{RIMES evaluation campaign for handwritten mail processing}},
  author={Augustin, Emmanuel and Carr{\'e}, Matthieu and Grosicki, Emmanu{\`e}le and Brodin, J-M and Geoffrois, Edouard and Pr{\^e}teux, Fran{\c{c}}oise},
  booktitle=iwfhr,
  year={2006}
}

@inproceedings{cascianelli2021learning,
  title={{Learning to Read L'Infinito: Handwritten Text Recognition with Synthetic Training Data}},
  author={Cascianelli, Silvia and Cornia, Marcella and Baraldi, Lorenzo and Piazzi, Maria Ludovica and Schiuma, Rosiana and Cucchiara, Rita},
  booktitle={CAIP},
  year={2021}
}

@inproceedings{quattrini2024alfie,
  title={{Alfie: Democratising RGBA Image Generation with No \$\$\$}},
  author={Quattrini, Fabio and Pippi, Vittorio and Cascianelli, Silvia and Cucchiara, Rita},
  booktitle=eccvw,
  pages={38--55},
  year={2024},
  organization={Springer}
}

@inproceedings{quattrini2024merging,
  title={Merging and splitting diffusion paths for semantically coherent panoramas},
  author={Quattrini, Fabio and Pippi, Vittorio and Cascianelli, Silvia and Cucchiara, Rita},
  booktitle=eccv,
  pages={234--251},
  year={2024},
  organization={Springer}
}

@inproceedings{quattrini2024mu,
  title={{$\mu$ gat: Improving Single-Page Document Parsing by Providing Multi-page Context}},
  author={Quattrini, Fabio and Zaccagnino, Carmine and Cascianelli, Silvia and Righi, Laura and Cucchiara, Rita},
  booktitle=eccvw,
  pages={212--228},
  year={2024},
  organization={Springer}
}

@article{pippi2023evaluating,
  title={{Evaluating Synthetic Pre-Training for Handwriting Processing Tasks}},
  author={Pippi, Vittorio and Cascianelli, Silvia and Baraldi, Lorenzo and Cucchiara, Rita},
  journal=prl,
  year={2023}
}

@inproceedings{quattrini2024binarizing,
  title={{Binarizing Documents by Leveraging Both Space and Frequency}},
  author={Quattrini, Fabio and Pippi, Vittorio and Cascianelli, Silvia and Cucchiara, Rita},
  booktitle=icdar,
  pages={3--22},
  year={2024},
  organization={Springer}
}

@article{vanherle2024vatr++,
  title={{VATr++: Choose Your Words Wisely for Handwritten Text Generation}},
  author={Vanherle, Bram and Pippi, Vittorio and Cascianelli, Silvia and Michiels, Nick and Van Reeth, Frank and Cucchiara, Rita},
  journal=ieeetpami,
  year={2024}
}

@inproceedings{dai2024one,
  title={{One-DM: One-Shot Diffusion Mimicker for Handwritten Text Generation}},
  author={Dai, Gang and Zhang, Yifan and Ke, Quhui and Guo, Qiangya and Huang, Shuangping},
  booktitle=eccv,
  year={2024}
}

@inproceedings{pippi2025zero,
  title={{Zero-Shot Styled Text Image Generation, but Make It Autoregressive}},
  author={Pippi, Vittorio and Quattrini, Fabio and Cascianelli, Silvia and Tonioni, Alessio and Cucchiara, Rita},
  booktitle=cvpr,
  year={2025}
}

@article{nikolaidou2024diffusionpen,
  title={{DiffusionPen: Towards Controlling the Style of Handwritten Text Generation}},
  author={Nikolaidou, Konstantina and Retsinas, George and Sfikas, Giorgos and Liwicki, Marcus},
  journal=eccv,
  year={2024}
}

@inproceedings{rombach2022high,
  title={{High-resolution image synthesis with latent diffusion models}},
  author={Rombach, Robin and Blattmann, Andreas and Lorenz, Dominik and Esser, Patrick and Ommer, Bj{\"o}rn},
  booktitle=cvpr,
  year={2022}
}

@article{mayr2024zero,
  title={{Zero-Shot Paragraph-level Handwriting Imitation with Latent Diffusion Models}},
  author={Mayr, Martin and Dreier, Marcel and Kordon, Florian and Seuret, Mathias and Z{\"o}llner, Jochen and Wu, Fei and Maier, Andreas and Christlein, Vincent},
  journal={arXiv preprint arXiv:2409.00786},
  year={2024}
}

@article{luhman2020diffusion,
  title={{Diffusion Models for Handwriting Generation}},
  author={Luhman, Troy and Luhman, Eric},
  journal={arXiv preprint arXiv:2011.06704},
  year={2020}
}

@inproceedings{ren2023diff,
  title={{Diff-Writer: A Diffusion Model-Based Stylized Online Handwritten Chinese Character Generator}},
  author={Ren, Min-Si and Zhang, Yan-Ming and Wang, Qiu-Feng and Yin, Fei and Liu, Cheng-Lin},
  booktitle=iconip,
  year={2023},
}

@article{raffel2020exploring,
  title={{Exploring the Limits of Transfer Learning with a Unified Text-to-Text Transformer}},
  author={Raffel, Colin and Shazeer, Noam and Roberts, Adam and Lee, Katherine and Narang, Sharan and Matena, Michael and Zhou, Yanqi and Li, Wei and Liu, Peter J},
  journal=jmlr,
  pages={1--67},
  year={2020}
}

@article{xue2022byt5,
    title = "{B}y{T}5: Towards a Token-Free Future with Pre-trained Byte-to-Byte Models",
    author = "Xue, Linting  and Barua, Aditya  and Constant, Noah and Al-Rfou, Rami  and Narang, Sharan  and Kale, Mihir  and Roberts, Adam  and Raffel, Colin",
    journal=tacl,
    pages = "291--306",
    year = "2022",
}

@misc{zhou2024transfusion,
  title={{Transfusion: Predict the next token and diffuse images with one multi-modal model}},
  author={Zhou, Chunting and Yu, Lili and Babu, Arun and Tirumala, Kushal and Yasunaga, Michihiro and Shamis, Leonid and Kahn, Jacob and Ma, Xuezhe and Zettlemoyer, Luke and Levy, Omer},
  booktitle=iclr,
  year={2025}
}

@misc{Infinity,
    title={{Infinity: Scaling Bitwise AutoRegressive Modeling for High-Resolution Image Synthesis}}, 
    author={Jian Han and Jinlai Liu and Yi Jiang and Bin Yan and Yuqi Zhang and Zehuan Yuan and Bingyue Peng and Xiaobing Liu},
    year={2025},
    booktitle=cvpr,
}

@article{fan2025unified,
  title={{Unified autoregressive visual generation and understanding with continuous tokens}},
  author={Fan, Lijie and Tang, Luming and Qin, Siyang and Li, Tianhong and Yang, Xuan and Qiao, Siyuan and Steiner, Andreas and Sun, Chen and Li, Yuanzhen and Zhu, Tao and others},
  journal={arXiv preprint arXiv:2503.13436},
  year={2025}
}

@inproceedings{
li2024autoregressive,
title={{Autoregressive Image Generation without Vector Quantization}},
author={Tianhong Li and Yonglong Tian and He Li and Mingyang Deng and Kaiming He},
booktitle=nips,
year={2024},
}

@inproceedings{
fan2025fluid,
title={{Fluid: Scaling Autoregressive Text-to-image Generative Models with Continuous Tokens}},
author={Lijie Fan and Tianhong Li and Siyang Qin and Yuanzhen Li and Chen Sun and Michael Rubinstein and Deqing Sun and Kaiming He and Yonglong Tian},
booktitle=iclr,
year={2025},
}

@article{team2025nextstep,
  title={{NextStep-1: Toward Autoregressive Image Generation with Continuous Tokens at Scale}},
  author={Team, NextStep and Han, Chunrui and Li, Guopeng and Wu, Jingwei and Sun, Quan and Cai, Yan and Peng, Yuang and Ge, Zheng and Zhou, Deyu and Tang, Haomiao and others},
  journal={arXiv preprint arXiv:2508.10711},
  year={2025}
}

@inproceedings{ho2021classifier,
  title={{Classifier-Free Diffusion Guidance}},
  author={Ho, Jonathan and Salimans, Tim},
  booktitle={NeurIPS 2021 Workshop on Deep Generative Models and Downstream Applications},
year={2021}
}

@article{nikolaidou2025dual,
  title={{Dual Orthogonal Guidance for Robust Diffusion-based Handwritten Text Generation}},
  author={Nikolaidou, Konstantina and Retsinas, George and Sfikas, Giorgos and Cascianelli, Silvia and Cucchiara, Rita and Liwicki, Marcus},
  journal=iccvw,
  year={2025}
}

@inproceedings{pippi2025quo,
  title={{Quo Vadis Handwritten Text Generation for Handwritten Text Recognition?}},
  author={Pippi, Vittorio and Nikolaidou, Konstantina and Cascianelli, Silvia and Retsinas, George and Sfikas, Giorgos and Cucchiara, Rita and Liwicki, Marcus},
  booktitle=iccvw,
  pages={7473--7483},
  year={2025}
}

@article{brandenbusch2024semi,
  title={{Semi-Supervised Adaptation of Diffusion Models for Handwritten Text Generation}},
  author={Brandenbusch, Kai},
  journal={arXiv preprint arXiv:2412.15853},
  year={2024}
}

@article{phunyaphibarn2025unconditionalpriorsmatterimproving,
            title={{Unconditional Priors Matter! Improving Conditional Generation of Fine-Tuned Diffusion Models}}, 
            author={Prin Phunyaphibarn and Phillip Y. Lee and Jaihoon Kim and Minhyuk Sung},
            year={2025},
            journal={arXiv preprint arXiv:2503.20240},
      }

@inproceedings{mu2025editar,
    author = {Mu, Jiteng and Vasconcelos, Nuno and Wang, Xiaolong},
    title = {EditAR: Unified Conditional Generation with Autoregressive Models},
    booktitle = cvpr,
    year = {2025}
}

@misc{chen2025context,
      title={Context-Aware Autoregressive Models for Multi-Conditional Image Generation}, 
      author={Yixiao Chen and Zhiyuan Ma and Guoli Jia and Che Jiang and Jianjun Li and Bowen Zhou},
      year={2025},
      eprint={2505.12274},
      archivePrefix={arXiv},
      primaryClass={cs.CV},
      url={https://arxiv.org/abs/2505.12274}, 
}

@article{liu2024glyph,
  title={Glyph-byt5: A customized text encoder for accurate visual text rendering},
  author={Liu, Zeyu and Liang, Weicong and Liang, Zhanhao and Luo, Chong and Li, Ji and Huang, Gao and Yuan, Yuhui},
  journal={arXiv preprint arXiv:2403.09622},
  year={2024}
}

@article{liu2024glyphv2,
  title={Glyph-ByT5-v2: A Strong Aesthetic Baseline for Accurate Multilingual Visual Text Rendering},
  author={Liu, Zeyu and Liang, Weicong and Zhao, Yiming and Chen, Bohan and Li, Ji and Yuan, Yuhui},
  journal={arXiv preprint arXiv:2406.10208},
  year={2024}
}

@article{tuo2023anytext,
      title={AnyText: Multilingual Visual Text Generation And Editing}, 
      author={Yuxiang Tuo and Wangmeng Xiang and Jun-Yan He and Yifeng Geng and Xuansong Xie},
      year={2023},
      eprint={2311.03054},
      archivePrefix={arXiv},
      primaryClass={cs.CV}
}

@misc{tuo2024anytext2,
      title={AnyText2: Visual Text Generation and Editing With Customizable Attributes}, 
      author={Yuxiang Tuo and Yifeng Geng and Liefeng Bo},
      year={2024},
      eprint={2411.15245},
      archivePrefix={arXiv},
      primaryClass={cs.CV},
}
}
\clearpage
\setcounter{page}{1}
\setcounter{section}{0}

\maketitlesupplementary

In this document, we report additional analyses on style text independence and the effect of the training strategies adopted in the second phase of training. Moreover, we report results that include color correction of the output.

\section{Style Text Reliance Analysis}
\label{sec:trocr_style_text}

When compared to Emuru~\cite{pippi2025zero}, Eruku does not need style text input. A possible workaround to use Eruku with a style sample with no known ground-truth textual transcription is to use an OCR model to obtain it. We test both Emuru and Eruku with style text input obtained from running TrOCR-Base~\cite{li2021trocr} as an OCR model and comparing them against each other when using the $T_s^*$ text generated by TrOCR, against Eruku ran with no style text input and against a version of Eruku which has never been trained with style text dropout. The results are displayed in~\Cref{tab:sup_style}.
Eruku is (except for FID) better than Emuru even when using the ground truth text $T_s$. When using $T^*_s$, Eruku is able to maintain very low $\Delta$CER, whereas Emuru tends to collapse and/or generate incorrect text more often. Both manage to maintain style consistency. Eruku with no style text gets even better $\Delta$CER scores, but compromises in a significant way on style adherence, as indicated by the high HWD score. The version of Eruku trained with no style text dropout and style text from OCR suffers, just like Emuru, from significantly increased $\Delta$CER from the reliance on this noisy style text.
Emuru is incapable of running with no style text input.

\section{Ablation on Second Stage Training}

In the second stage of pretraining, as described in ~\Cref{sec:training}, two variations are made to the way the model trains: it is trained on the dataset of images with longer context described in~\Cref{ssec:dataset}, and it is trained to randomly drop style text conditioning with a probability of $p_{drop}=0.1$. We investigate the effects of each of those by running training for the same amount of iterations as the full Eruku second stage of training, but with just one strategy or the other. We then compare those runs on IAM lines to the full second stage of training and to the result of just the first stage of training.
The results, shown in~\Cref{tab:sup_training}, highlight how long-context training improves $\Delta$CER significantly. Style text dropout instead, in addition to allowing the model to generate unconditionally as shown in~\Cref{sec:experiments}, also improves style image adherence, as indicated by the improvement in HWD. The model using both strategies (Eruku) combines the advantages of both and reaches the best HWD values and much-improved $\Delta$CER values when compared to the model resulting from the first stage of training.

\section{Results Including Color Correction}
Since it relies on the same VAE as Emuru, Eruku generates images with a white background and usually very dark text strokes. This allows the simple color correction strategy proposed in~\cite{pippi2025zero} for Emuru to be applicable also for Eruku. The strategy uses the VAE's background removal abilities to isolate the mask containing the text within the style image, then computes the average of the color values among the foreground ink pixels and applies that to those of the generated image. 
The effect of such color correction post-processing can be observed quantitatively in~\Cref{tab:sup_cc} (especially in terms of FID).

\begin{table}[t]
    \centering
    \setlength{\tabcolsep}{.7em}
    \resizebox{\columnwidth}{!}{%
    \begin{tabular}{l cccc}
    \toprule
    & \textbf{HWD$\downarrow$}
    & \textbf{$\Delta$CER$\downarrow$}
    & \textbf{FID$\downarrow$}
    & \textbf{BFID$\downarrow$} \\
    \midrule
   \textbf{Eruku w/ $T_s$}   &  \textbf{1.70} & \textbf{0.06} & 16.40 & \textbf{4.88} \\
   \textbf{Emuru w/ $T_s$}   & 1.87 & 0.14 & \textbf{13.89} & 6.19 \\
   \midrule
   \textbf{Eruku w/ $T_s^*$} & 1.73 & \textbf{0.06} & 16.59 & \textbf{5.07} \\
   \textbf{Eruku $p_{drop}=0$ w/ $T_s^*$} & \textbf{1.72} & 0.53 & 15.81 & 7.68 \\
   \textbf{Emuru w/ $T_s^*$} & 1.79 & 0.42 & \textbf{14.09} & 6.23 \\
   \midrule
   \textbf{Eruku w/o $T_s$}  & \textbf{2.51} & \textbf{0.04} & \textbf{20.44} & \textbf{9.63} \\
   \textbf{Emuru w/o $T_s$}  & - & - & - & - \\
    \bottomrule
    \end{tabular}}
\caption{Emuru and Eruku results on IAM lines when fed with the actual $T_s$ or a $T_s$ obtained by running TrOCR on the style image $I_s$ (dubbed $T_s^*$). As a reference, we report the results of the generation without $T_s$.}\vspace{-0cm}\label{tab:sup_style}
\end{table}

\begin{table}[t]
    \centering
    \setlength{\tabcolsep}{.3em}
    \resizebox{\columnwidth}{!}{%
    \begin{tabular}{c c cccc}
    \toprule
    \textbf{longer input}
    & $p_{drop} = 0.1$
    & \textbf{HWD$\downarrow$}
    & \textbf{$\Delta$CER$\downarrow$}
    & \textbf{FID$\downarrow$}
    & \textbf{BFID$\downarrow$} \\
    \midrule
    \xmark & \xmark & 1.81 & 0.40 & 14.20 & \textbf{3.38} \\
    \checkmark & \xmark & 1.92 & \textbf{0.04} & 19.45 & 5.50 \\
    \xmark & \checkmark & 1.75 & 0.40 & \textbf{13.49} & 4.45 \\
     \checkmark & \checkmark & \textbf{1.70} & 0.06 & 16.40 & 4.88 \\
    \bottomrule
    \end{tabular}
}\caption{Ablation analysis on the effect of the second training phase inputs and strategy in terms of performance on IAM Lines.}\vspace{-0cm}
\label{tab:sup_training}
\end{table}

\begin{table}[t]
    \centering
    \setlength{\tabcolsep}{.7em}
    \resizebox{\columnwidth}{!}{%
    \begin{tabular}{l cccc}
    \toprule
    & \textbf{HWD$\downarrow$}
    & \textbf{$\Delta$CER$\downarrow$}
    & \textbf{FID$\downarrow$}
    & \textbf{BFID$\downarrow$} \\
    \midrule
   \textbf{Eruku}   &  1.70 & 0.06 & 16.40 & 4.88 \\
   \textbf{Emuru}   & 1.87 & 0.14 & 13.89 & 6.19 \\
   \textbf{Eruku w/ c.c.} & \textbf{1.68} & \textbf{0.04} & 12.21 & \textbf{4.54} \\
   \textbf{Emuru w/ c.c.} & 1.85 & 0.14 & \textbf{11.40} & 6.20 \\
    \bottomrule
    \end{tabular}}
\caption{Emuru and Eruku results on IAM lines in the standard setting and when the color correction strategy (c.c.) is applied as post-processing.}\vspace{-0cm}\label{tab:sup_cc}
\end{table}

\end{document}